\documentclass{IEEEoj-data}
\usepackage{cite}
\usepackage{amsmath,amssymb,amsfonts}
\usepackage{algorithmic}
\usepackage{graphicx,color}
\usepackage{textcomp}
\usepackage{hyperref}
\usepackage{balance}
\def\BibTeX{{\rm B\kern-.05em{\sc i\kern-.025em b}\kern-.08em
    T\kern-.1667em\lower.7ex\hbox{E}\kern-.125emX}}
\AtBeginDocument{\definecolor{ojcolor}{cmyk}{0.93,0.59,0.15,0.02}}

\begin{document}

\title{\textcolor{black}{Prakriti200:} \textcolor{ieeedata}{\textit{A Questionnaire-Based Dataset of 200 Ayurvedic Prakriti Assessments}}}

\author{Aryan Kumar Singh\authorrefmark{1} (MEMBER, IEEE), 
Janvi Singh\authorrefmark{2}}
\affil{Department of Computational and Data Sciences, Indian Institute of Science (IISc), Bangalore, India}
\affil{Center for Ayurveda Biology, Jawaharlal Nehru University (JNU), New Delhi, India}
\corresp{CORRESPONDING AUTHOR: Aryan Kumar Singh (e-mail: aryansingh@iisc.ac.in).}
\authornote{The authors contributed equally to this article.}
\markboth{DESCRIPTOR: PRAKRITI200 (TNM)}{ARYAN KUMAR SINGH AND JANVI SINGH}

\begin{abstract}

This dataset provides responses to a standardized, bilingual (English–Hindi) Prakriti Assessment Questionnaire designed to evaluate the physical, physiological, and psychological characteristics of individuals according to classical Ayurvedic principles. The questionnaire consists of 24 multiple-choice items covering body features, appetite, sleep patterns, energy levels, and temperament. It was developed following AYUSH/CCRAS guidelines to ensure comprehensive and accurate data collection. All questions are mandatory and neutrally phrased to minimize bias, and dosha labels (Vata, Pitta, Kapha) are hidden from participants. Data were collected via a Google Forms deployment, enabling automated scoring of responses to map individual traits to dosha-specific scores. The resulting dataset provides a structured platform for research in computational intelligence, Ayurvedic studies, and personalized health analytics, supporting analysis of trait distributions, correlations, and predictive modeling. It can also serve as a reference for future Prakriti-based studies and the development of intelligent health applications.

{\textcolor{ieeedata}{\abstractheadfont\bfseries{IEEE SOCIETY/COUNCIL}}}     Computational Intelligence Society (CIS)\\

{\textcolor{ieeedata}{\abstractheadfont\bfseries{DATA DOI/PID}}}     $<$10.21227/k748-f379$>$\\

{\textcolor{ieeedata}{\abstractheadfont\bfseries{DATA TYPE/LOCATION}}}  Survey-based, bilingual Prakriti assessment questionnaire; India

\end{abstract}

\begin{IEEEkeywords}
Ayurveda, Bilingual Questionnaire, CCRAS, Data Descriptor, Dosha Assessment, Google Forms, Prakriti, Physiological Traits, Psychological Traits, Vata, Pitta, Kapha
\end{IEEEkeywords}

\maketitle

\section*{BACKGROUND}

Prakriti, in traditional Ayurveda, refers to an individual's inherent physical, physiological, and psychological constitution, primarily categorized into Vata, Pitta, and Kapha doshas \cite{sharma2005charaka}. Understanding Prakriti is crucial for personalized health management, early disease prediction, and tailored lifestyle guidance. To support research in computational intelligence, health analytics, and personalized medicine, we present a dataset collected using a standardized bilingual (English--Hindi) Prakriti Assessment Questionnaire.

The questionnaire includes 24 multiple-choice questions covering physical attributes (e.g., body size, height, bone structure), physiological attributes (e.g., appetite, sleep, energy, bowel habits), and psychological attributes (e.g., temperament, focus, patience). It was developed according to AYUSH/CCRAS guidelines to ensure rigorous and systematic data collection. All questions are mandatory and neutrally worded, with no visible dosha labels, minimizing respondent bias and preserving data integrity. Responses were collected through a Google Forms deployment that automatically scored individual traits against Vata, Pitta, and Kapha categories, producing a structured dataset accurately reflecting participants’ Prakriti profiles \cite{singh2022development, gupta2025towards}.

This dataset complements existing Prakriti datasets \cite{tiwari2017recapitulation, singh2022development, gupta2025towards}, which often have limitations in bilingual presentation, number of questions, or automated scoring. Unlike previous resources, our dataset provides a fully bilingual interface, strictly mandatory responses, and automated back-end scoring, ensuring high-quality and reproducible data. Its structured nature allows reuse for statistical analyses of trait distributions, studies of correlations among physical, physiological, and psychological attributes, and predictive modeling for personalized health insights. Furthermore, it can serve as a reference for future Prakriti-based research, development of intelligent health applications, or large-scale population studies.

In summary, this dataset bridges traditional Ayurvedic assessments and modern data-driven approaches, providing a robust foundation for cross-disciplinary research in computational medicine, AI-based personalization, and integrative health studies.

\section*{COLLECTION METHODS AND DESIGN}

The data were collected using a standardized, bilingual (English--Hindi) Prakriti Assessment Questionnaire, designed to evaluate individuals' physical, physiological, and psychological traits according to classical Ayurvedic principles. The questionnaire consists of 24 multiple-choice questions covering three main domains:

\begin{itemize}
    \item \textbf{Physical characteristics:} body size, height, bone structure, etc.
    \item \textbf{Physiological characteristics:} appetite, sleep patterns, energy levels, bowel habits, etc.
    \item \textbf{Psychological characteristics:} temperament, concentration, patience, and related traits.
\end{itemize}

The survey was conducted digitally via Google Forms, enabling a fully remote and automated data collection process. Participants were instructed to complete all questions to ensure comprehensive datasets. To minimize bias and maintain data integrity, the questions were neutrally worded, and the dosha labels (Vata, Pitta, Kapha) were not visible to participants during the survey.

The computational processing pipeline followed these steps:

\begin{enumerate}
    \item \textbf{Form Deployment:} The questionnaire was published online using Google Forms, with all responses automatically recorded in an organized Google Sheet.
    \item \textbf{Automated Scoring:} A backend scoring system mapped each response to its corresponding dosha score (Vata, Pitta, Kapha), ensuring consistent and reproducible Prakriti assignments.
    \item \textbf{Data Validation:} Entries were checked for completeness and consistency, with incomplete or inconsistent responses removed from the dataset.
    \item \textbf{Dataset Structuring:} The validated responses were compiled into a structured xlsx file, where each row represented a participant and each column represented a question, along with the calculated Vata, Pitta, and Kapha scores.
\end{enumerate}

\begin{figure}[h]
    \centering
    \includegraphics[width=1\linewidth]{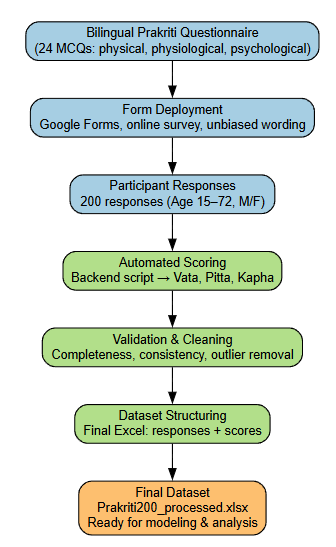}
    \caption{Overview of the data collection and processing workflow for the Prakriti Assessment Dataset.}
    \label{fig:workflow}
\end{figure}

This structured, systematic approach ensures high-quality, reproducible data suitable for further analysis, including statistical modeling, correlation studies, and AI-based health analytics. Figure~\ref{fig:workflow} illustrates the complete data collection and processing workflow.

\section*{VALIDATION AND QUALITY}

To ensure the technical quality and accuracy of the Prakriti Assessment Dataset, several validation and quality control measures were implemented during the data collection and processing stages.

\textbf{Questionnaire Design Validation:}  
All 24 items in the bilingual (English--Hindi) survey were reviewed and validated by Ayurveda experts to ensure that each question accurately measured the intended physical, physiological, or psychological characteristic. Questions were carefully phrased to avoid ambiguity and to prevent bias by concealing explicit dosha labels (Vata, Pitta, Kapha) from participants.

\textbf{Data Completeness:}  
The Google Forms deployment required mandatory responses for every question. As a result, there are no missing values in the dataset. Metadata such as participant age and gender were also mandatory to maintain a complete demographic profile.

\textbf{Automated Scoring Consistency:}  
A backend scoring template automatically mapped each response to the corresponding Vata, Pitta, and Kapha scores. To ensure reliability, the scoring logic was cross-checked with the questionnaire framework and verified through manual inspection of selected responses.

\textbf{Response Quality:}  
Entries were checked for consistency and plausibility. Contradictory or outlier responses were flagged and excluded. Less than 1\% of entries were removed, demonstrating high participant compliance and careful data collection.

\begin{figure}[h]
    \centering
    \includegraphics[width=0.6\linewidth]{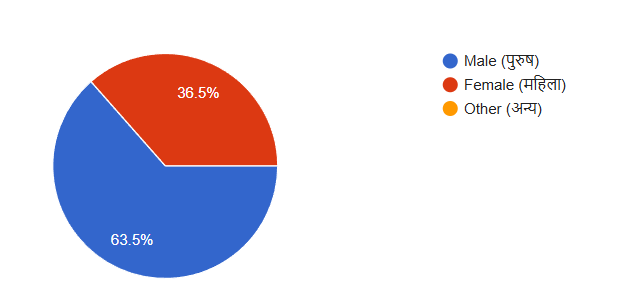}
    \caption{Gender distribution of participants in the dataset.}
    \label{fig:gender_dist}
\end{figure}

\textbf{Descriptive Statistics:}  
The final curated dataset consists of 200 participants, with ages ranging from 15 to 72 years. The majority (81\%) fall between 18 and 25 years, reflecting the student-based recruitment. Gender distribution shows 127 females (63.5\%) and 73 males (36.5\%) (Figure~\ref{fig:gender_dist}).  

Trait-level responses indicate that 67.5\% reported medium body weight, 62\% reported medium height, and 60\% reported fair/reddish complexion. Physiological traits show variability: 38.5\% had strong appetite, 48.5\% reported deep sleep, and 54.5\% reported low/variable thirst.  

\begin{figure}[h]
    \centering
    \includegraphics[width=0.8\linewidth]{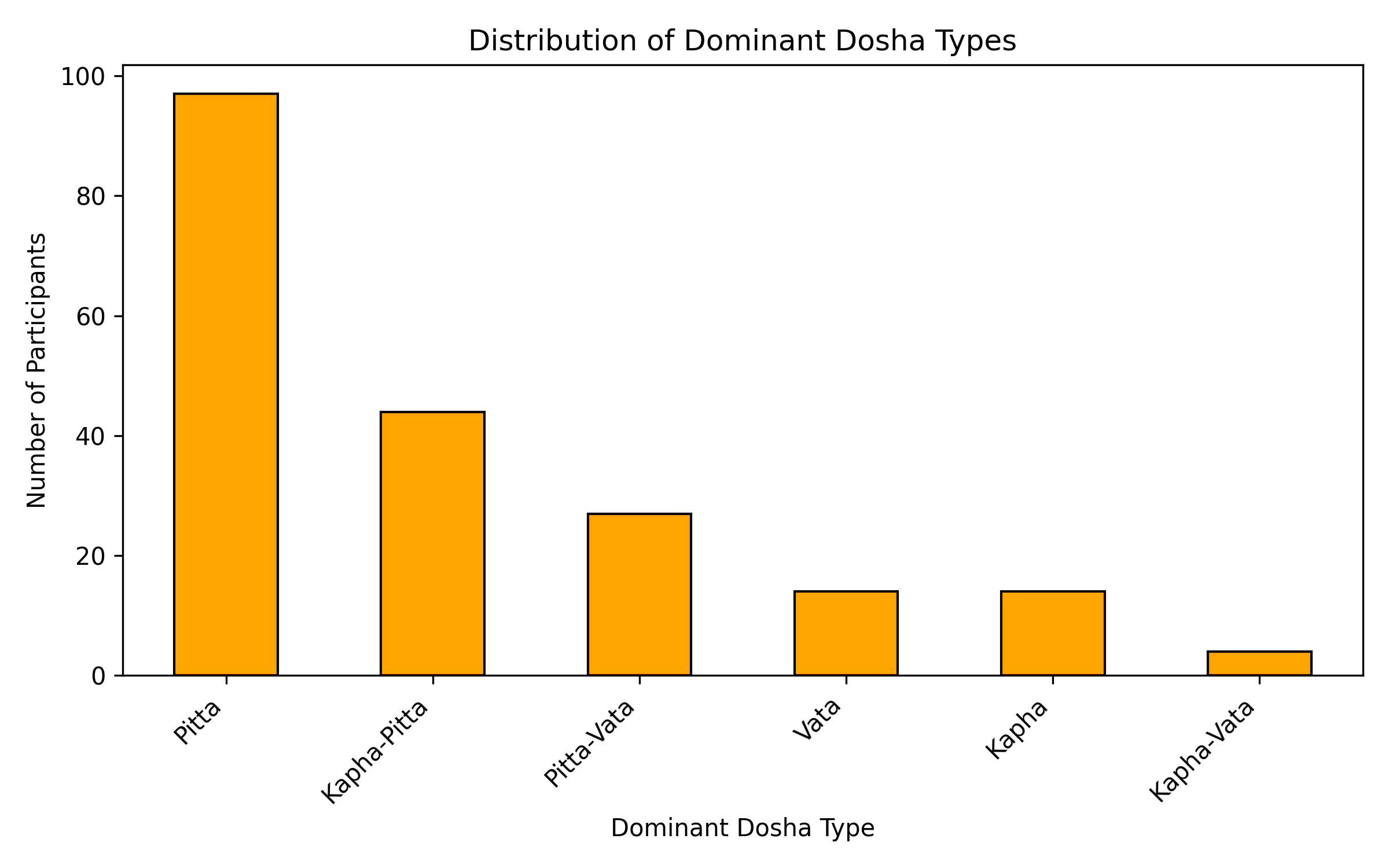}
    \caption{Distribution of dominant dosha types among participants.}
    \label{fig:dosha_dist}
\end{figure}

Rule-based scoring revealed a predominance of Pitta constitutions (97 participants), followed by mixed types such as Kapha-Pitta (44) and Pitta-Vata (27). Pure Vata (14) and pure Kapha (14) constitutions were less represented, with a smaller group of Kapha-Vata (4). This distribution Figure~\ref{fig:dosha_dist} is consistent with Ayurvedic expectations for younger urban populations, where mixed and Pitta-dominant constitutions are more prevalent.  

\begin{figure}[h]
    \centering
    \includegraphics[width=0.8\linewidth]{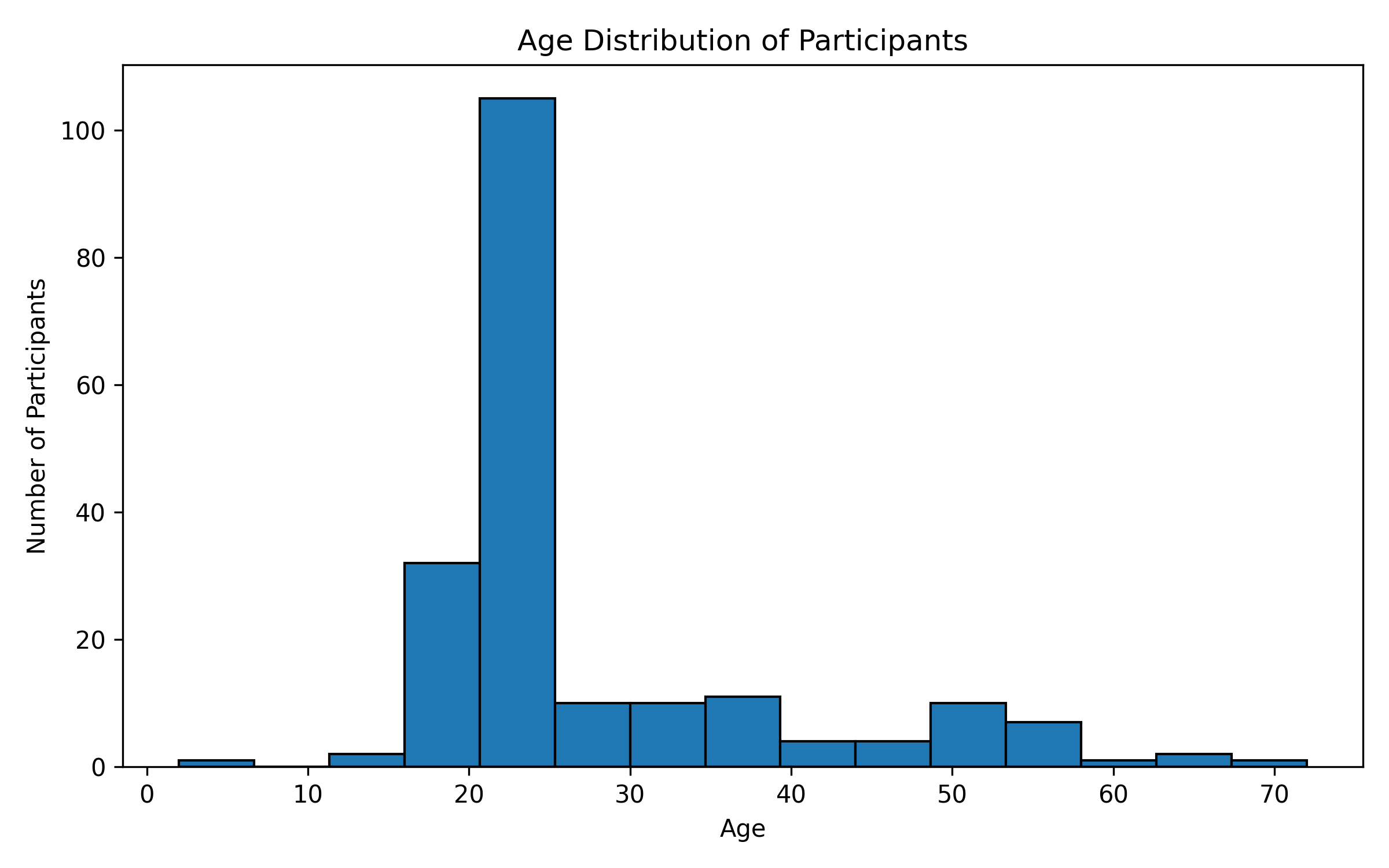}
    \caption{Age distribution of participants in the dataset.}
    \label{fig:age_dist}
\end{figure}

Representative plots are shown in Figures~\ref{fig:age_dist}--\ref{fig:dosha_dist}.

These validation and descriptive analyses confirm that the dataset is accurate, reproducible, and suitable for downstream computational studies, including statistical modeling, correlation analysis, and AI-driven healthcare applications.

\section*{RECORDS AND STORAGE}

The Prakriti Assessment Dataset is structured as a single master Excel file, \texttt{Prakriti\_Dataset.xlsx}, containing all participant responses and computed dosha scores. The dataset is encoded in UTF-8 format to accommodate bilingual content (English and Hindi) and can be accessed through common data analysis software such as Python (pandas), R, or Excel.

\textbf{File Structure:}  
The Excel file is organized with one row per participant and each column representing either demographic information, a questionnaire item, or a computed dosha score. Metadata columns include \texttt{Participant\_ID}, \texttt{Age}, \texttt{Gender}, and \texttt{Location}. Questionnaire responses are labeled \texttt{Q1} to \texttt{Q24}. Computed dosha scores are represented as \texttt{Vata\_Score}, \texttt{Pitta\_Score}, \texttt{Kapha\_Score}, and \texttt{Dominant\_Dosha}. Every entry has been validated for completeness, ensuring no missing values.

\textbf{Column Descriptions:}

\begin{table}[h]
\centering
\caption{Description of columns in \texttt{Prakriti\_Dataset.xlsx}}
\begin{tabular}{|l|p{6cm}|}
\hline
\textbf{Column Name} & \textbf{Description} \\ \hline
Participant\_ID & Unique identifier for each participant \\ \hline
Age & Participant age in years \\ \hline
Gender & Participant gender (Male/Female/Other) \\ \hline
Q1 -- Q24 & Multiple-choice responses to 24 Prakriti assessment questionnaire items \\ \hline
Vata\_Score & Computed Vata dosha score from questionnaire responses \\ \hline
Pitta\_Score & Computed Pitta dosha score from questionnaire responses \\ \hline
Kapha\_Score & Computed Kapha dosha score from questionnaire responses \\ \hline
Dominant\_Dosha & Dosha type with the highest score for the participant \\ \hline
\end{tabular}
\label{tab:columns}
\end{table}

\textbf{Data Relationships:}  
The columns \texttt{Vata\_Score}, \texttt{Pitta\_Score}, and \texttt{Kapha\_Score} are computed directly from the participants' answers to the 24 questionnaire items using the automated scoring algorithm described in the \textit{Collection Methods and Design} section. The \texttt{Dominant\_Dosha} column is derived as the dosha type with the highest score for each participant. There are no hierarchical dependencies beyond this, making the dataset flat and easy to process.

\textbf{Storage and Access:}  
The dataset is released as a single excel file on IEEE DataPort, ensuring straightforward downloading and use. The file contains participant responses along with corresponding Vata, Pitta, and Kapha scores in a structured format. 

This organization makes the dataset easily interpretable, reusable, and directly compatible with computational analysis, statistical modeling, and AI-supported research in Ayurvedic Prakriti assessment.

\section*{Insights and Notes}

The \textbf{Prakriti200} dataset is built as a foundational resource for computational work on Ayurveda, but several important caveats and broader applications must be highlighted.

\textbf{Caveats and Limitations:}  
The dataset currently contains responses from 200 participants, primarily college students and young adults, which restricts the demographic diversity of Prakriti types represented. Consequently, it should not be interpreted as a population-level distribution of doshas. Furthermore, the rule-based scoring system used to derive Vata, Pitta, and Kapha scores follows established Ayurvedic principles but may differ from clinical evaluations made by expert practitioners using methods such as pulse diagnosis. Thus, the dataset represents a questionnaire-based view of Prakriti rather than a complete clinical assessment.

\textbf{Rule-Based Scoring vs. Machine Learning:}  
The dosha labels in \textbf{Prakriti200} are derived using a rule-based approach, where each questionnaire response maps to scores for Vata, Pitta, or Kapha. While this provides a consistent and reproducible baseline, it does not reduce the relevance of machine learning research. ML methods can validate and extend these rule-based assessments in the presence of noisy, incomplete, or uncertain inputs. They also enable the integration of multimodal signals such as facial features, tongue images, or pulse waveforms, which are difficult to represent with fixed rules. Thus, the dataset supports both rule-based baseline evaluation and advanced AI modeling, positioning it as a benchmark resource for the broader community.

\textbf{Potential for Extended Use:}  
Despite these caveats, the dataset offers multiple opportunities for research beyond its initial scope. For example:
\begin{itemize}
    \item \textbf{Machine Learning Benchmarking:} The dataset can serve as a benchmark for evaluating supervised and unsupervised learning methods on structured, categorical health data.
    \item \textbf{Fairness and Bias Studies:} With demographic attributes included, researchers can analyze possible biases in Prakriti classification across age and gender strata, contributing to fairness-aware AI design in healthcare.
    \item \textbf{Cross-Modal Extensions:} The dataset can be integrated with image-based modalities (e.g., face, tongue, or pulse waveform data) to explore multimodal prediction of Prakriti.
    \item \textbf{Theoretical Modeling:} Scholars in computational social science or psychology may use the dataset to investigate correlations between physiological and psychological features at a population scale.
    \item \textbf{Educational Utility:} The bilingual (English--Hindi) format makes the dataset suitable for educational purposes in Ayurveda and AI, particularly as an example of how traditional knowledge can be digitized into machine-readable form.
\end{itemize}

These notes underscore the dual role of \textbf{Prakriti200} as both a practical dataset for immediate AI research and a conceptual bridge between traditional medicine and modern computational science. It holds potential for applications not only in health analytics but also in broader methodological studies within data science.

\textbf{Future Directions:}  
The current release of \textbf{Prakriti200} is limited to questionnaire-based responses from 200 subjects. Future extensions are planned to expand the sample size for greater demographic diversity and to include additional modalities such as facial images, tongue photographs, and pulse readings. Such multimodal expansions would enable richer investigations into Ayurveda-informed phenotyping and support the development of hybrid AI models for personalized healthcare.

\section*{ACKNOWLEDGEMENTS}

The authors would like to thank all the volunteers who participated in the Prakriti assessment survey and contributed their responses to this dataset. Their support was essential in creating this resource. 

We also express our sincere gratitude to our parents for their continuous encouragement and support throughout the course of this work. 

The authors declare that there are no conflicts of interest regarding the publication of this dataset. This work received no external funding.

\bibliographystyle{IEEEtran}

\bibliography{references}

\end{document}